\documentclass{article}

\usepackage{aaai}

\copyrighttext{}

\newcommand{\Smodels}{\textsc{Smodels}}
\newcommand{\smodels}{\mbox{{\tt smodels}}}
\newcommand{\lparse}{\mbox{{\tt lparse}}}

\begin{document}

\title{\Smodels: A System for Answer Set Programming\thanks{%
The work has been funded by the Academy of Finland
(Project 43963) and by 
Helsinki Graduate School in Computer Science and Engineering. 
}} 

\author{Ilkka Niemel\"a \and Patrik Simons \and Tommi Syrj\"anen \\
Helsinki University of Technology \\ 
Dept.\ of Computer Science and Eng. \\
Laboratory for Theoretical Computer Science, \\
P.O.Box 5400, FIN-02015 HUT, Finland \\
\{Ilkka.Niemela, Patrik.Simons, Tommi.Syrjanen\}\@hut.fi
}

\maketitle

\begin{abstract}
\noindent The \Smodels\ system implements the stable model semantics for normal
logic programs. It handles a subclass of programs which contain no
function symbols and are domain-restricted but supports extensions including 
built-in functions as well as cardinality and weight
constraints. 
On top of this core engine  more involved systems can be built. As an
example, we have implemented total and partial stable model computation for
disjunctive logic programs.
An interesting application method is based on answer set programming, i.e.,
encoding an application problem as a set of rules so that its solutions
are captured by the stable models of the rules. 
\Smodels\ has been applied to a number of areas including 
planning, model checking, reachability analysis, 
product configuration, dynamic constraint satisfaction, and feature
interaction.
\end{abstract}

\section{General Information}


The \Smodels\ system is written in C++ and the source code, test cases
and documentation are available at
\texttt{http://www.tcs.hut.fi/Software/smodels/}. In order to compile the system
a C++ compiler is needed as well as other standard tools such
as make, tar, bison, and perl. The system has been developed under Linux and should
work as is on any platform having the appropriate GNU tools installed.
It has been used on a wide range of hardware (PC/SPARC/Alpha) mostly
running Unix.
The total number of lines of codes is about 20000.

\section{Description of the System} 

The \Smodels\ system implements the stable model semantics for normal
logic programs extended by built-in functions as well as cardinality
and weight constraints for domain-restricted programs. In this section
we briefly discuss the syntax, implementation techniques and use of
the system. More information can be found at the home page
\texttt{http://www.tcs.hut.fi/Software/smodels/}.

As input the \Smodels\ system takes logic program rules basically in 
Prolog style syntax:
\begin{verbatim}
    ancestor(X,Y) :- ancestor(X,Z), 
                     parent(Z,Y),  
                     person(X). 
    ancestor(X,Y) :- parent(X,Y). 
    son(X,Y) :- parent(Y,X), male(X). 
    daughter(X,Y) :- parent(Y,X), female(X). 
    person(X) :- male(X). 
    person(X) :- female(X). 
    parent(jack, jill). parent(joan, jack). 
    male(jack). female(jill). female(joan). 
\end{verbatim}

However, in order to support efficient implementation techniques and
extensions the programs are required to be \emph{domain-restricted}
where the idea is the following. No proper function symbols are
allowed (but we do allow built-in functions) and the predicate symbols
in the program are divided into two classes, \emph{domain predicates}
and \emph{non-domain predicates}. 
Domain predicates are predicates that are defined
non-recursively. In the program above all predicates except
\verb+ancestor+ are domain predicates. The predicate \verb+ancestor+
is not a domain predicate because it depends recursively on itself. 

The main intuition of domain predicates is that they are used to define 
the set of terms over which the variables range in each rule of a program $P$.
All rules of $P$ have to be domain-restricted in the sense that every
variable in a rule must appear in a domain predicate which appears
positively in the rule body. For instance, in the first rule of the program
above all variables appear in domain predicates \verb+parent+ and
\verb+person+ in the body of the rule.

In addition to normal logic program rules, \Smodels\  supports rules with
cardinality and weight constraints. 
The idea is that, e.g.,  a cardinality constraint
\[
\verb+1 { a,b,not c } 2+
\]
holds in a stable model if at the least 1 but at most 2 of the literals
in the constraint are satisfied in the model and a weight constraint
\[
\verb+ 10 [ a=10, b=10, not c=10 ] 20+
\]
holds if the sum of weights of the literals satisfied in the model is
between 10 and 20 (inclusive).
With built-in functions for integer arithmetic (included in the
system), these kinds of rules allow compact and fairly straightforward
encodings of many interesting problems. For example, the N queens
problem can be captured using rules as a program \verb+queens.lp+ as follows:
\begin{verbatim}
    1 { q(X,Y):d(X) } 1 :- d(Y).
    1 { q(X,Y):d(Y) } 1 :- d(X).
    :- d(X), d(Y), d(X1), d(Y1),
       q(X,Y), q(X1,Y1),
       X != X1,  Y != Y1, 
       abs(X - X1) == abs(Y - Y1).
    d(1..n).
\end{verbatim}
where \verb+d(1..n)+ is a domain predicate giving the dimension of the
board (from 1 to integer $n$ which can be specified during run time). 
The first rule says that for each row $y$, a stable model contains
exactly one atom $q(x,y)$ where for $x$, $d(x)$ holds and similarly
in the second rule for all columns.
The third rule is an integrity constraint saying that there cannot be
two queens on the same diagonal.
Now each stable model corresponds to a legal configuration of $n$ queens
on a $n\times n$ board, i.e., $q(x,y)$ is in a stable 
model iff $(x,y)$ is a legal position for a queen.

Stable models of a domain-restricted logic program with variables are
computed in three stages. First, the program is transformed into
a ground program without variables. Second, the rules of the
ground program are translated into primitive rules, and third, a
stable model is computed using a Davis-Putnam like
procedure~\cite{Simons99:lpnmr}. The first two stages have been implemented
in the program \lparse, which functions as a front end
to \smodels\ which in turn implements the third stage.

In the first stage \lparse\ automatically determines the domain
predicates and then using database techniques evaluates the domain
predicates and creates a ground program which has exactly the same
stable models as the original program with variables. 
Then the rules are compiled into primitive rules~\cite{NSS99:lpnmr}.

The \smodels\ procedure is a Davis-Putnam like backtracking search procedure
that finds the stable models of a set of primitive rules by assigning truth values to
the atoms of the program. Moreover, it uses the properties of the
stable model semantics to infer and propagate additional truth
values. Since the procedure is in effect traversing a binary search
tree, the number of nodes in the search space is in the worst case on
the order of $2^n$, where $n$ can be taken to be the number of atoms
that appear in a constraint in a head of a rule or that appear as a
negative literal in a recursive loop of the program.

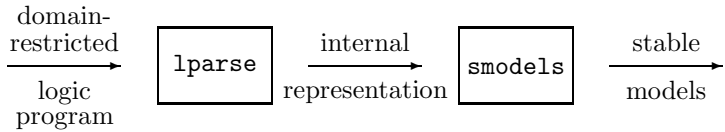
\begin{figure*}
\setlength{\unitlength}{1mm}
\center
\begin{picture}(100,15)(0,0)
\put(0,10){\vector(1,0){15}}
\put(7.5,12){\makebox(0,0)[b]{\shortstack{domain-\\ restricted}}}
\put(7.5,8){\makebox(0,0)[t]{\shortstack{logic\\ program}}}
\put(20,5){\framebox(15,10){\lparse}}
\put(40,10){\vector(1,0){15}}
\put(47.5,12){\makebox(0,0)[b]{internal}}
\put(47.5,8){\makebox(0,0)[t]{representation}}
\put(60,5){\framebox(15,10){\smodels}}
\put(80,10){\vector(1,0){15}}
\put(87.5,12){\makebox(0,0)[b]{stable}}
\put(87.5,8){\makebox(0,0)[t]{models}}
\end{picture}
\caption{Overall Architecture}\label{fig:architecture}
\end{figure*}

Hence, in order to compute stable models, one uses the two programs
\lparse, which translates logic
programs into an internal format, and \smodels, which
computes the models, see 
Figure~\ref{fig:architecture}. 

For instance, a solution to the 8 queens problem given the program 
\verb+queens.lp+ above, would be typically computed by a command line:
\begin{verbatim}
lparse -c n=8 -d none queens.lp | smodels
\end{verbatim}
where \verb+-c n=8+ option instructs to use the value 8 for the constant $n$
and \verb+-d none+ option instructs to remove all domain predicates from
the rules as soon as they have been evaluated. The command line produces output:
\begin{verbatim}
Answer: 1
Stable Model: q(4,1) q(2,2) q(7,3) q(5,4) 
q(1,5) q(8,6) q(6,7) q(3,8) 
\end{verbatim}

\subsection{A Further Example}

The graph coloring problem may be encoded to a program
\verb+ncolor.lp+ using the following two \Smodels\ rules:
\begin{verbatim}
  1 { col(N, C) : color(C) } 1 :- node(N).
  :- col(X, C), col(Y,C), edge(X,Y), color(C).
\end{verbatim}
Here the predicate \verb+col(N,C)+ denotes that the color of the node
\verb+N+ is \verb+C+. The first rule states that each node has exactly
one color and the second rule states that two adjacent nodes may not
have the same color. Suppose we have a fully connected three node
graph and we want to find all 3-colorings of it where the first node is
colored red. We can encode the problem instance to a program
\verb+graph.lp+:
\begin{verbatim}
  node(a ; b; c).
  edge(a,b). edge(a,c). edge(b,c).
  color(red ; green ; blue).
  compute { col(a, red) }.
\end{verbatim}
The first two lines define the graph and the third line defines the
three colors. The \verb+compute+ statement tells \Smodels\ that we are
interested only in those models where \verb+col(a, red)+ is true, i.e., the node \verb+a+ has color \verb+red+. We can find all stable
models that satisfy the compute statement with the command line:
\begin{verbatim}
lparse -d none ncolor.lp graph.lp | smodels 0
\end{verbatim}
This input produces the output:
\begin{verbatim}
Answer: 1
Stable Model: col(c,blue) col(b,green)
col(a,red) 
Answer: 2
Stable Model: col(c,green) col(b,blue)
col(a,red) 
\end{verbatim}

More information about the syntax and use of the system can be found in
the \lparse\ user's manual at
\texttt{http://www.tcs.hut.fi/Software/smodels/lparse/} and about the 
implementation in~\cite{Simons99:lpnmr,NSS99:lpnmr}.

\section{Applying the System}


\subsection{Methodology}


An interesting application methodology for \Smodels\ is based on answer set
programming~\cite{MT99:slp,Niemela99:amai} which has emerged as a viable
approach to declarative logic-based knowledge representation.  It is
based on the stable model semantics of logic programs and can be seen as
a novel form of constraint programming where constraints are expressed
as rules.  The underlying idea is to encode an application problem using
logic program rules so that the solutions to the problem are captured by
the stable models of the rules. The solution of the N queens problem in
the previous section illustrates nicely main ideas of answer set
programming.


\subsection{Specifics}


It is important that the system is based on an
implementation-independent declarative semantics. This makes it much
easier to develop applications because one does not have to worry too
much about the internal implementation-specific aspects of the system.
Hence, the system is relatively easy to learn to use. On the other
hand, declarative semantics provides much more flexibility in
developing implementation approaches and in optimizing different parts
of the implementation. We have taken advantage of this and developed
methods which are substantially different from usual implementation
methods for logic programming (Prolog) but still work efficiently in
new kinds of applications where Prolog style systems are not
appropriate.

\Smodels\ implements the stable model semantics for range-restricted
function free normal programs. It can also compute well-founded models
for these programs.
\Smodels\ supports built-in functions, e.g.,
for integer arithmetic. Basic \Smodels\ extends normal logic programs
with cardinality and weight
constraints~\cite{Simons99:lpnmr,NSS99:lpnmr}. On top of this core engine
 more involved systems can be built. As an example, we have implemented
total and partial stable model computation for disjunctive logic
programs~\cite{JNSY2000:kr}.

The semantics for logic programs with cardinality and weight constraints
supported by the core engine is an interesting compromise: it is rather
simple to learn, its complexity stays in NP in the ground case (like for
propositional logic) but it seems strictly more expressive than
propositional logic (or other standard constraint satisfaction
formalisms).

With our work on \Smodels\ we hope to demonstrate that nonmonotonic
reasoning techniques are useful conceptual tools as well as bring
computational advantages which can lead to new interesting applications
and to developments of novel implementation techniques and tools.

\subsection{Users and Usability} 
The basic semantics of \Smodels\ rules seems to be simple
enough that it can be explained to a non-expert or a student in a
relatively short amount of time sufficiently so that one can start using
\Smodels. 

\Smodels\ has been employed in a number of areas including 
planning~\cite{DNK97,Niemela99:amai,Lifschitz99:iclp}, 
model checking~\cite{LRS98:tacas}, reachability 
analysis~\cite{Heljanko99:tacas,Heljanko:Fundamenta99}, 
product configuration~\cite{SN99:padl,Syrjanen99:dtyo}, 
dynamic constraint satisfaction~\cite{SNG99:cp}, feature
interaction~\cite{AAR99}, and logical cryptanalysis~\cite{HMN2000:nmr}.

In order to make \Smodels\ more flexible an API has been added to
\lparse\ and \smodels. Hence, they can be used through the
API and embedded into a C/C++ program as 
libraries. Furthermore, it is possible to define new built-in functions
for the front-end \lparse\ in order to accommodate new applications.
These new functions are written in C/C++ and they are dynamically
linked to \lparse\ when needed. 

As a further usability feature, \lparse\ performs simple analysis of
the program and warns about constructs that are often erroneous. For
example, \lparse\ detects if a variable is accidentally mistyped as a
constant or vice versa. These warnings can be enabled with command
line options.

\section{Evaluating the System}


\subsection{Benchmarks}

We have compiled quite a large collection of families of benchmarks that
we use to evaluate new developments and improvements in our system and
which we can employ to compare our system to other competing ones 
and which can be used also by others
(see, \texttt{http://www.tcs.hut.fi/Software/smodels/tests/}).
Similar testing methodology (e.g., generating test
cases from graph problems) has also been used by the groups in Kentucky
(DeReS/TheoryBase system~\cite{CMMT99:aij})  and in
Vienna  (dlv system~\cite{ELMPS98}). 
These kinds of tests can be used for measuring the base level
performance and to compare different systems. 
However, it is unclear what is a portable way of representing such
benchmarks. We use Prolog style syntax which is quite generally
accepted but it seems that each system has its specialties. 


\subsection{Comparison} 

\Smodels\ can already compete with special purpose systems and 
we have even cases where it outperforms commercial top edge tools, e.g., 
in a verification application it has performance better than that of 
one of the most efficient commercial integer programming tools (CPLEX)
\cite{Heljanko:Fundamenta99}.

\subsection{Problem Size} 

We believe that \Smodels\ is no longer a mere prototype but it can
handle realistic size problems. We have applications where 
programs with hundreds of thousands of non-trivial ground rules are
treated efficiently, see e.g.~\cite{Niemela99:amai,Heljanko:Fundamenta99}. 



\end{document}